\pdfoutput=1
\documentclass[11pt]{article}
\usepackage[final]{acl}
\usepackage{times}
\usepackage[T1]{fontenc}
\usepackage[utf8]{inputenc}
\usepackage{graphicx}
\usepackage{booktabs}
\usepackage{colortbl}
\definecolor{tablehead}{HTML}{EAF0F8}   %
\definecolor{tablehl}{HTML}{FBECEA}     %
\usepackage{amsmath}
\usepackage{amssymb}
\usepackage{xcolor}

\newcommand{\sys}{ARGUS}
\usepackage{framed}
\definecolor{shadecolor}{gray}{0.95}
\newcommand{\promptlabel}[1]{\par\addvspace{4pt}\noindent\textbf{#1}\par\nobreak\vspace{1pt}}
\newenvironment{promptbox}%
  {\par\addvspace{2pt}\begin{shaded}\footnotesize\ttfamily\raggedright\noindent}%
  {\end{shaded}\addvspace{2pt}}

\title{Evidence-Grounded Auditing of Identification Assumptions\\
in Climate-Policy Causal Evaluations}

\author{
  Yonghong Zhang$^{1}$ \quad Yong Xie$^{2}$ \quad
  Isabel M. Parra$^{1}$ \quad Ricardo Correia$^{1}$ \\
  $^{1}$Department of Finance, Universidad Aut\'onoma de Madrid, Madrid, Spain \\
  $^{2}$Spanish National Research Council (CSIC), Madrid, Spain \\
  Correspondence: \texttt{yonghong.zhang@estudiante.uam.es}
}

\hypersetup{pdftitle={Evidence-Grounded Auditing of Identification Assumptions in Climate-Policy Causal Evaluations},
  pdfauthor={Yonghong Zhang, Yong Xie, Isabel M. Parra, Ricardo Correia}}
\begin{document}
\maketitle

\begin{abstract}
Difference-in-differences (DID) studies are widely used to evaluate climate
policy, but assessing the evidence supporting their identification assumptions
remains challenging. We introduce \sys{}, a structured language-model pipeline
that audits reported evidence against an eleven-dimension
assumption--implication--evidence rubric and abstains when relevant evidence
cannot be retrieved. We evaluate \sys{} using injected flaws, economics
papers, and a small pilot with reconciled labels. On the 11-flaw benchmark,
\sys{} detects $73\%$ of planted flaws, compared with $18\%$ for a keyword-based
pipeline. Across 26 economics papers, \sys{} abstains on about $40\%$ of
paper--dimension assessments for lack of retrievable evidence. In a five-paper
pilot with labels reconciled by two annotators, it assigns a higher risk level
than the labels on 25 of the 33 assessments it completes. A rule fixed before
the labels arrived removes most of this in-sample; weighted agreement stays low. \sys{} provides evidence-linked risk reports
that localize potential weaknesses for expert review, without adjudicating
causal claims. Code and data:
\url{https://github.com/yonghongzhang-io/ARGUS}.
\end{abstract}

\section{Introduction}
\label{sec:intro}

Difference-in-differences (DID) is widely used to evaluate climate and
environmental policy, including carbon pricing, emissions-trading schemes, and
emission standards. The credibility of these estimates depends on identification
assumptions such as parallel trends, no anticipation, appropriate treatment of
staggered adoption, and limited interference between units, each of which a large
body of methodological work has shown to be fragile
\cite{bertrand2004trust,goodmanbacon2021,callaway2021,sunabraham2021,dechaisemartin2020,roth2023}.
These assumptions can be hard to assess in climate settings, where policies may
overlap, be phased in over time, spill across administrative boundaries, and
rely on modelled or remotely sensed outcomes (Appendix~\ref{app:motivation}).
Where such estimates inform policy, causal conclusions built on weakly
supported assumptions can mislead.

Assessing an identifying assumption requires more than locating a reported
diagnostic. A paper may present parallel trends over an event-study figure,
leaving the reader to weigh the design, the pre-period estimates, the
inference, and the policy timing before accepting the claim; the assessment
has to be repeated for every assumption and every paper. This is one facet of a
broader concern with the reliability of empirical findings
\cite{ioannidis2005,silberzahn2018many,brodeur2020methods}.

\begin{figure*}[t]
  \centering
  \includegraphics[width=\textwidth]{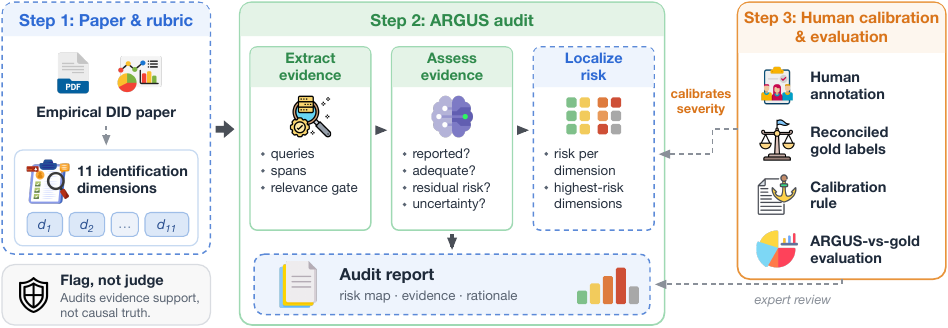}
  \caption{Overview of \sys{}. Step~1 converts an empirical DID paper into an
  auditable structure using an 11-dimension identification rubric. Step~2
  retrieves dimension-specific evidence, assesses reported-evidence adequacy with
  a bounded LLM audit module, and localizes risk in a human-reviewable report.
  Step~3 uses labels from two annotators, reconciled by them, to evaluate
  audit severity and score a pre-specified calibration rule offline; deployment
  requires only expert review of the report. Colours as in Figure~\ref{fig:arch}: blue, deterministic steps;
  green, model calls; orange, human steps.}
  \label{fig:overview}
\end{figure*}

We argue that identification credibility is auditable even when the true causal
effect is not, building on the research-design tradition in applied econometrics
\cite{angristpischke2010,atheyimbens2017,rambachanroth2023}. Because the
counterfactual is unobserved, \sys{} (named after Argus Panoptes, the
hundred-eyed watchman) does not ask whether an estimated effect is true. Instead, it asks whether each identifying
assumption is adequately supported by the evidence reported in the paper, and where
that support is weakest. \sys{} represents a DID study through an 11-dimension
\emph{assumption--implication--evidence} rubric and uses a bounded,
retrieval-gated pipeline built on a large language model (LLM) to retrieve
relevant evidence, assess its adequacy, and produce an evidence-grounded risk
report for expert review (Figure~\ref{fig:overview}). Model agency is confined
to scoped retrieval and assessment tasks behind a fixed, deterministic control flow,
unlike free-roaming agents whose control flow is model-decided \cite{yao2023react};
\sys{} flags potential weaknesses rather than adjudicating causal truth.

Evaluating such an auditor is difficult because the true policy effect is
unobserved. We therefore use \emph{flaw injection} as local ground truth: starting
from a supported study fixture, we introduce one known identification flaw and test
whether \sys{} detects it, remains quiet on the clean version, and localizes the
affected dimension. We complement this controlled evaluation with a run over 26 corpus papers
tagged DID and a small diagnostic pilot with reconciled labels, used to
identify calibration and applicability failures. Climate-policy evaluation
motivates the rubric; the present evaluation uses synthetic
environmental-policy fixtures and a general-economics corpus, so performance on
a dedicated climate-policy corpus is untested.

\paragraph{Contributions.}
We develop and evaluate a rubric and pipeline for assessing the reported
evidence behind DID identification assumptions, without access to the causal
effect:
\begin{enumerate}
  \item an \emph{operationalization}: a machine-auditable
  \emph{assumption--implication--evidence} rubric and a bounded,
  evidence-grounded pipeline with explicit abstention (\S\ref{sec:framework});
  \item an \emph{evaluation methodology}: planted identification errors as
  local ground truth, with compute-graded and oracle ablations that locate the
  gain in per-dimension prompting and trace every synthetic omission miss to a
  gate abstention (\S\ref{sec:results});
  \item a \emph{diagnosis}: on real papers, grounding limits \emph{coverage}
  while over-severity limits \emph{agreement}, and a pre-specified
  rule derived on the same papers reduces the latter in-sample.
\end{enumerate}

\section{Related Work}
\label{sec:related}

\paragraph{DID identification and its threats.}
A large econometrics literature formalizes when DID estimators fail under
staggered timing and heterogeneous effects: two-way fixed effects' use of
already-treated units as controls~\cite{goodmanbacon2021}, negative
weights~\cite{dechaisemartin2020}, robust
estimators~\cite{callaway2021,sunabraham2021}, assumption and pre-trend
surveys~\cite{roth2023}, and fragile inference under serial
correlation~\cite{bertrand2004trust}. \sys{} adds no estimator: it
\emph{operationalizes} this literature as an auditable rubric, turning each
known threat into a dimension whose supporting evidence can be checked in a
given paper (\S\ref{sec:framework}).

\paragraph{LLM causal reasoning and benchmarks.}
A parallel line asks whether LLMs reason about causality. Benchmarks
(CLadder~\cite{jin2023cladder}, Corr2Cause~\cite{jin2024corr2cause},
CausalBN-Bench~\cite{zhou2024causalbench}) and
analyses~\cite{kiciman2023causal,yang2024criticalreview,zecevic2023causalparrots}
report mixed evidence: strong performance on some causal tasks, but models
often reproduce causal \emph{language} without reliable causal
\emph{inference}; a few move toward applied settings and toward separating
\emph{identification} from
\emph{estimation}~\cite{lee2025econcausal,sawarni2026causalreasoning}. \sys{}
addresses a different task: whether a \emph{human} paper's identification
assumptions are adequately evidenced. A causal verdict is out of scope by design.

\paragraph{Responsible AI for empirical research.}
Artificial intelligence (AI) is increasingly applied to research reliability: reproducibility checking,
statistical-reporting validation, and AI-assisted review, motivated by wide
analyst-driven result spreads~\cite{silberzahn2018many,botviniknezer2020variability}
and specification searching in causal economics~\cite{brodeur2020methods}. LLMs
are also used directly as evaluators and reviewers, judging model
outputs~\cite{zheng2023judge}, giving feedback on research
papers~\cite{liang2024feedback}, and reviewing inside automated-science
pipelines~\cite{lu2024aiscientist}; these evaluators already produce ratings,
pairwise preferences, and structured reviews. \sys{} applies an
assumption-specific rubric to DID studies, links each dimension to in-paper
evidence, and is measured against injected ground truth. \sys{}
focuses on the evidence for \emph{identification credibility} specifically
(the assumptions supporting the causal interpretation) and issues no verdicts, producing evidence-grounded, human-reviewable reports that localize
risk for the expert's final call. Closest to our setting,
CausalVerify~\cite{causalverify} grounds evaluation in \emph{execution}: it runs
a model's generated estimation code against synthetic data and checks the
recovered effect. That benchmark does not audit identification assumptions, which is the
layer \sys{} takes up; we reuse its released
corpus of economics papers and add a dimension-level
identification-risk annotation layer on top of it.

\paragraph{Natural language processing for climate evidence.}
Natural language processing (NLP) systems for climate text verify claims
against evidence
(CLIMATE-FEVER~\cite{climatefever2020}), detect environmental
claims~\cite{stammbach2023environmental}, analyse corporate climate disclosures
with traceable LLM answers~\cite{ni2023chatreport}, and tune models for
faithful evidence-based question answering~\cite{schimanski2024faithful}.
These systems ground answers in documents; \sys{} extends document-grounded
verification from \emph{claims} to the \emph{identification argument} of a
causal study, auditing whether reported evidence supports each assumption.

\paragraph{Evaluation methodology and bounded agents.}
Our evaluation follows benchmarking lessons: holistic
evaluation~\cite{liang2023helm} over a single number; dynamic, adversarial
benchmarking~\cite{kiela2021dynabench} (which inspires our controlled flaw injection);
and the caution that operationalizing a capability is
non-neutral~\cite{raji2021benchmark,gebru2021datasheets}
(\S\ref{sec:discussion}). A closely related concern is that an agent
benchmark can measure its harness rather than its model:
\citet{dmconfound} show that a fixed scaffold making the execution-critical
decisions and a scorer grading output shape mask each other, so repairing
either alone leaves the ranking uninformative. Our ablations apply a
related check inward: they compare joint and per-dimension prompting, with and
without retrieval and relevance gating, holding the model and rubric fixed
(\S\ref{sec:variants}). Where ReAct-style agents~\cite{yao2023react} choose
their own action sequence, \sys{} fixes the order of retrieval and assessment
and confines model calls to those two stages, trading autonomy for
inspectability and evidence-traceability.

\section{The \sys{} Framework}
\label{sec:framework}

\subsection{Identification rubric}
\sys{} decomposes a DID study into eleven auditable dimensions of
identification, design, inference, and reporting: parallel trends, no
anticipation, staggered-timing handling, SUTVA (stable unit treatment value
assumption, no interference) and spillovers,
control-group construction, specification, inference, sample-period selection,
concurrent policies, placebo/robustness, and data measurement. Each dimension is
expressed as an \emph{assumption} or design requirement (what must hold), a
testable \emph{implication} (the diagnostic that should then be observable),
and the \emph{evidence} a credible paper would report to support it. For a carbon-pricing
or emissions-trading evaluation, for instance, the parallel-trends dimension
expects an event-study or pre-trend test with pre-period estimates near zero
(an insignificant pre-trend alone can reflect low power; \citealp{roth2022}), while
no-anticipation turns on the gap between a scheme's \emph{announcement} and
\emph{implementation} dates. The rubric is declarative, stored in a configuration
file (\texttt{identification\_dimensions.yaml}), so the audited dimensions are
explicit and inspectable rather than buried in code. A companion flaw taxonomy
(\S\ref{sec:evaluation}) maps each known identification threat one-to-one onto a
dimension, which both documents the rubric's coverage and supplies the
perturbations used for evaluation.

\subsection{Bounded LLM pipeline}
\label{sec:pipeline}
\begin{figure*}[t]
  \centering
  \includegraphics[width=\textwidth]{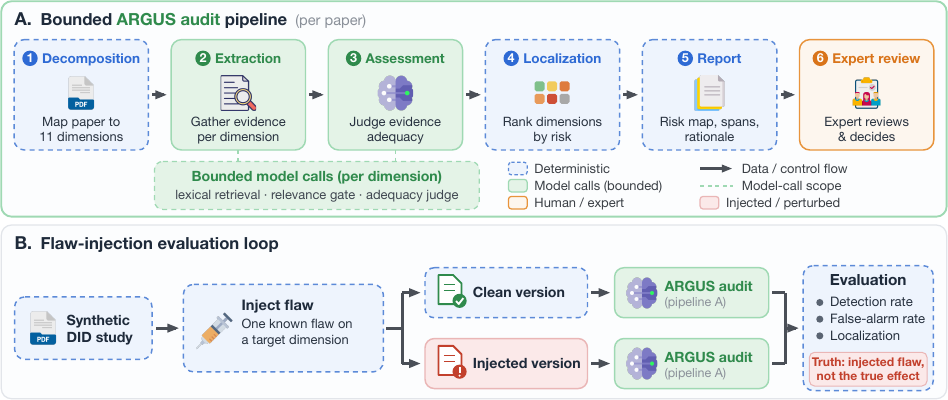}
  \caption{\sys{} architecture. The top band is the audit pipeline, a fixed
  deterministic sequence; model calls are confined to two stages (extraction
  and assessment). As evaluated, these resolve per dimension to a fixed call
  sequence (lexical retrieval, one relevance-gate call, one adequacy call) with
  a step budget of one and a logged trace; the designed tools
  \texttt{figure\_parse} and \texttt{policy\_lookup} are not exercised
  (\S\ref{sec:pipeline}). The bottom band shows the flaw-injection evaluation loop, which
  injects one known flaw, audits the clean and injected versions, and scores
  detection, false alarm, and localization against the injected flaw as local
  ground truth, never against the unobserved true effect.}
  \label{fig:arch}
\end{figure*}

The audit runs a fixed, deterministic sequence, \emph{decomposition
$\rightarrow$ extraction $\rightarrow$ assessment $\rightarrow$ localization
$\rightarrow$ report}, shown in the top band of Figure~\ref{fig:arch}. The stage
order is written in code and never chosen by a model. \emph{Decomposition}
instantiates the rubric for the paper at hand; \emph{extraction} gathers, per
dimension, the in-paper evidence relevant to that assumption; \emph{assessment}
evaluates each assumption--implication--evidence chain and emits a per-dimension
risk judgement with a rationale and cited evidence; \emph{localization}
aggregates these judgements into a risk map that ranks the dimensions assigned
the highest risk (ties broken by the fixed rubric order); and
\emph{report} renders a transparent, evidence-grounded document for a human
reviewer.

Model calls are confined to two stages, extraction and assessment. In the
configuration evaluated in \S\ref{sec:results} they resolve, for each
dimension, to a fixed sequence: deterministic lexical retrieval over the
paper's sections and captions, one relevance-gate call that labels each
retrieved passage \emph{high}, \emph{partial}, or \emph{none}, and, if at least
one passage is kept, one adequacy call. If no passage is kept the dimension is
scored \texttt{unknown} without an adequacy call; a dimension with only
partially relevant passages is still judged (Appendix~\ref{app:retrieval}).
Figures and tables enter only through their captions. The code also defines a
more general bounded loop for these two stages (a fixed tool set of
\texttt{evidence\_search}, \texttt{figure\_parse}, and
\texttt{policy\_lookup}; a hard step budget, \texttt{max\_steps}; a logged
trace of every step and cited passage); the reported experiments run it with
\texttt{max\_steps}{=}1 and never invoke \texttt{figure\_parse} or
\texttt{policy\_lookup}, and all other stages are plain orchestration with
no model call. We therefore call the design
\emph{bounded} rather than agentic: the evaluated system is structured,
dimension-scoped prompting over retrieved evidence with an explicit abstention
path, not an open-ended agent.

\paragraph{Why bound the model's role.}
Fixing the stage order makes the sequence of operations explicit and comparable
across runs and keeps every judgement anchored to logged evidence a human can
inspect. It does not make model outputs identical: Appendix~\ref{app:repro}
reports the variation observed across repeated runs. The expert sits at the end
of the pipeline: \sys{} surfaces and localizes identification risk; the
judgement itself stays with the human.

\subsection{Output}
\sys{} emits study--dimension--risk triplets, each with an
evidence-grounded rationale, alongside the human-readable report. In the
evaluated prompts (Appendix~\ref{app:prompts}) \emph{low}, \emph{medium},
and \emph{high} correspond approximately to retrieved evidence judged
sufficient, partial, and missing or flawed; \texttt{unknown} is produced by the
gate when no relevant passage is retained and reflects retrieval, not a
judgement about the paper. None of the four is a verdict on the causal effect. This
output shape keeps the system both auditable by a human and measurable by the
evaluation that follows.

\section{Evaluation by Flaw Injection}
\label{sec:evaluation}

\subsection{Local ground truth without the counterfactual}
The defining difficulty in evaluating a causal-identification auditor is that the
quantity of ultimate interest (the true causal effect of a carbon price, an
emissions cap, or an emission standard) is never observed. We therefore do not
attempt to certify that \sys{} recovers true effects. Instead we construct
\emph{local} ground truth by controlled perturbation. Beginning from a short
synthetic fixture, an environmental-policy DID study that reports supporting
evidence on all eleven dimensions, we inject
exactly one known flaw that targets a single dimension, drawn from the flaw
taxonomy (\texttt{flaw\_taxonomy.yaml}; Appendix~\ref{app:flawlit} traces
each flaw to the literature that names it). The target dimension is known by
construction, and the expected response (risk of at least \emph{medium} on
that dimension) is defined by the rubric, without ever needing the true
effect. The bottom band of Figure~\ref{fig:arch} shows the loop: the clean and
injected versions are each audited, and three metrics are computed.

\begin{itemize}
  \item \textbf{Detection}: does the injected version raise risk on the targeted
        dimension to \emph{medium} or above? (All reported results use this
        threshold.)
  \item \textbf{False alarm}: does the clean version stay quiet on that dimension?
  \item \textbf{Localization}: is the targeted dimension the one the auditor flags
        most strongly?
\end{itemize}

\subsection{Structural perturbation, not sentinels}
How a flaw is injected is a methodological choice that matters. A na\"ive injector
appends a sentence that \emph{names} the flaw: e.g.\ ``treated and control have
non-zero pre-period coefficients with significant leads.'' If the detector then
searches for exactly those phrases, detection becomes a tautology: the injector
plants the very string the detector matches on. In our initial implementation this
produced a detection rate of $1.000$ inflated by phrase leakage.

We replace this with \emph{structural perturbation} of two kinds.
\emph{Omission} flaws \emph{remove} the section, figure, or table that supplies
the evidence the rubric specifies for a dimension (e.g.\ deleting the
placebo-robustness section or the balance table). They test sensitivity to
missing reported support; they do not establish that the underlying assumption
is violated. \emph{Commission} flaws \emph{rewrite} a section to introduce a
stated design problem in prose intended to resemble how a flawed paper would
describe it (e.g.\ pooling all province-years in a two-way fixed-effects
regression under staggered policy timing), \emph{without} the phrases the
keyword detector looks for. A regression test fails if an injected paper
contains the target dimension's predefined negative-signal phrases, so this
phrase leakage cannot silently return; detection cannot come from matching
those phrases.

\section{Results}
\label{sec:results}

We evaluate \sys{} along four axes: a synthetic flaw-injection protocol, an
architecture ablation, a run over 26 papers tagged DID in an economics corpus,
which stress-tests the audit architecture across empirical economics beyond
the climate-policy setting that defines its target deployment, and a pilot with reconciled labels. Four
findings emerge: (i)~LLM pipelines that judge evidence \emph{adequacy} detect
planted flaws that a keyword-presence pipeline misses; (ii)~per-dimension
prompting detects more planted flaws than a single joint call, while the
retrieval gate turns retrieval failure into an auditable \texttt{unknown}
signal; (iii)~on real papers retrieval limits coverage, and weakly relevant
passages are associated with high-risk ratings; and (iv)~against the
reconciled labels \sys{} is systematically over-severe, a bias that a
pre-specified rule derived on the same papers reduces in-sample.

\subsection{Adequacy pipelines versus keyword presence; what the gate trades}
\label{sec:ablation}

We compare three assessors under one flaw-injection protocol on the
\texttt{clean\_supported} fixture (one flaw injected per run, all eleven flaws;
\texttt{max\_steps}{=}1). A deterministic \textbf{keyword baseline} scores each
dimension from the presence of supporting and risk-indicating vocabulary. The
\textbf{two-stage \sys{} assessor} judges evidence \emph{adequacy} (whether the
reported evidence supports the assumption under the rubric). The two are complete
pipelines with their own evidence paths: the baseline scores paragraph chunks,
while the two-stage path re-retrieves and gates whole sections
(Appendix~\ref{app:retrieval}). The pair is therefore an \emph{end-to-end}
comparison and cannot separate the judgement policy from retrieval.
A \textbf{single-pass} ablation removes the architecture entirely: the
full paper and all eleven dimensions in one \texttt{gpt-4o} call, with the same
model, temperature (the sampling-randomness setting, held at $0$), risk schema,
and scoring harness, but no staged retrieval,
relevance gating, or bounded tool use; its strict output schema admits only
low, medium, or high, so this arm cannot abstain
(Appendix~\ref{app:prompts}). Table~\ref{tab:results} reports the headline metrics.

\begin{table}[t]
  \centering\small
  \caption{Synthetic flaw injection on the \texttt{clean\_supported} fixture, eleven flaws;
  all LLM rows gpt-4o, detection replicated on the dated 2024-11-20 snapshot; see
  Appendix~\ref{app:repro} on run-to-run differences). Every row is an end-to-end pipeline with its own evidence path (keyword and
  two-stage do not share retrieved evidence, Appendix~\ref{app:retrieval};
  Appendix~\ref{app:shared} holds it fixed); the
  ablation arms share the model and rubric but remove retrieval, the gate, or
  both. The final row is the leaked, pre-fix
  circular setup. On the 33-variant benchmark the two-stage assessor
  rates one clean dimension medium, a $0.09$ over-flag (\S\ref{sec:variants}).
  Wilson 95\% intervals for the main rates are in Table~\ref{tab:ci}.}
  \label{tab:results}
  \setlength{\tabcolsep}{5pt}
  \resizebox{\columnwidth}{!}{%
  \begin{tabular}{lcccc}
    \toprule
    \rowcolor{tablehead} assessor & calls & detection & false alarm & localization \\
    \midrule
    keyword baseline             & 0  & 0.182 & 0.000 & 0.182 \\
    single-pass                  & 1  & 0.818 & 0.000 & 0.818 \\
    per-dimension, no retrieval  & 11 & 1.000 & 0.000 & 1.000 \\
    \rowcolor{tablehl} \textbf{two-stage \sys{}}    & 22 & \textbf{0.727} & \textbf{0.000} & \textbf{0.727} \\
    \midrule
    \textit{(circular, pre-fix)} & -- & 1.000 & 0.000 & 1.000 \\
    \bottomrule
  \end{tabular}}
\end{table}

\begin{figure*}[t]
  \centering
  \includegraphics[width=0.72\textwidth]{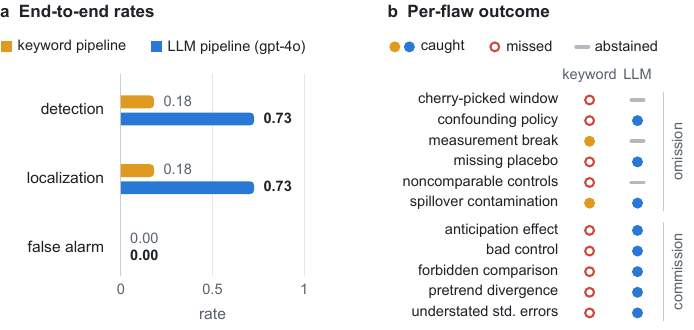}
  \caption{Keyword pipeline vs LLM evidence-adequacy pipeline ($n{=}11$ flaws,
  structural injection). \textbf{(a)}~End to end, the LLM pipeline raises detection and
  localization $0.182\,{\to}\,0.727$ with false alarm $0$ (the two pipelines
  retrieve evidence differently; Appendix~\ref{app:shared} holds it fixed). \textbf{(b)}~Per-flaw
  caught/missed: the keyword baseline is \emph{blind to every commission-type
  flaw}, all of which the LLM catches; the LLM's three misses are omissions on
  which it abstains. Table~\ref{tab:variants} reports the 33-variant
  benchmark.}
  \label{fig:results}
\end{figure*}

\paragraph{The keyword baseline is blind to commission flaws.}
The baseline never raises a false alarm and correctly flags the two omission
flaws (\texttt{measurement\_break} and \texttt{spillover\_contamination}) by the
\emph{absence} of their vocabulary, but beyond those detection is only $0.182$
(Figure~\ref{fig:results}b). Two failure modes explain it: the same keywords occur in
other sections (deleting the placebo section does not help when another
section also mentions ``placebo'') and blindness to \emph{every} commission
flaw, whose flawed-but-plausible evidence carries the expected topic vocabulary.

\paragraph{Detection improves with the LLM pipelines.}
Replacing the keyword pipeline with the two-stage LLM pipeline raises detection
from $0.182$ to $0.727$ with no false alarms, catching \emph{every} commission
flaw the keyword scorer is blind to; handing each dimension the full fixture (the
per-dimension arm) reaches $1.000$. Since retrieval also differs, a
pre-specified control holds the evidence fixed (Appendix~\ref{app:shared}).
Reading the keyword pipeline's own evidence chunks, with no gate, the adequacy
judge detects 10 of 11 flaws against the keyword scorer's 2 (exact McNemar
$p{=}0.008$), and 31 against 8 of the 33 variants. The gain is a trade: the
judge rates one clean dimension medium and $11$--$16\%$ of non-target
dimensions on injected papers, where the keyword scorer flags none. The model reads flawed-but-natural prose the keyword scorer
cannot; the audit report in Appendix~\ref{app:qual} shows it flagging the
injected anticipation flaw with the announcement-gap rationale, with no sentinel
keyword present.

\paragraph{What the gate trades.}
The gated pipeline detects \emph{less} than its own ablations here ($0.727$ vs
$0.818$ and $1.000$) by design: its three misses are all \emph{omission} flaws
on which retrieval finds nothing and it abstains rather than guessing; in the
reported runs each of these misses terminates at the relevance gate as
\texttt{unknown}, and the oracle-retrieval arm of \S\ref{sec:variants} flags
the same cases when the assessor is called on the target section. The gate's effect is behavioural and shows on real papers. A single-pass
variant whose prompt \emph{does} permit \texttt{unknown} (``use sparingly'';
Appendix~\ref{app:prompts}) abstained on one of 275 cells ($0.4\%$), against
$38.9\%$ mechanical abstentions for the gated pipeline
(Appendix~\ref{app:matrix}). The designs behave differently when evidence is
hard to ground; without cell-level gold the contrast does not show which is
more reliable.

\paragraph{Scope.}
These are short synthetic fixtures; ceilings on clearly injected flaws are a
proof-of-concept, tested next on 33 variants (\S\ref{sec:variants}) and real
papers (\S\ref{sec:realpapers}).

\subsection{An expanded benchmark of 33 flaw variants}
\label{sec:variants}
The eleven flaws are deliberately clear-cut, so we built an expanded benchmark:
33 variants, three per dimension (two \emph{commission} rewrites, one
\emph{omission}). The rewrites were written to resemble descriptions of flawed
design choices and are verified free of the keyword detector's predefined
negative-signal phrases. Three two-stage runs (gpt-4o, temperature~$0$) are near-deterministic
(detection identical on 32/33 variants, risk labels on 31/33; Appendix~\ref{app:repro}). Detection is $0.73$--$0.76$ and localization
$0.64$--$0.70$, in line with its behaviour on the eleven clear flaws.

\paragraph{Where the gain comes from.}
A compute-graded ablation on this benchmark (Appendix~\ref{app:matrix})
removes retrieval and gating in turn, the de-scaffolding check
of~\citet{dmconfound} applied to our own pipeline: a full-paper single pass
(one call) detects $0.727$; per-dimension prompting without retrieval (eleven
calls) detects $0.879$; the full two-stage pipeline (twenty-two calls) detects
$0.747$. Per-dimension prompting detects more variants than a single joint
call (29/33 vs 24/33); the comparison changes both the task decomposition and
the number of calls, so it does not isolate their separate contributions. The
retrieval-and-gate layer then \emph{trades} omission detection for abstention
(on these short fixtures a gateless assessor sees the deleted section in its
context; on real papers it commits judgements nearly everywhere,
\S\ref{sec:realpapers}). An \emph{oracle-retrieval} arm supplies
a diagnostic comparison: fed the target section directly (empty for omissions,
bypassing the gate), the assessor detects $0.86$ of commissions and $1.00$ of
omissions, while the gate suppresses the alarms section-level evidence alone
triggers ($0.21$ oracle arm vs $0.09$; the full-context arm's $0.00$ reflects
its wider view). At $n{=}33$ none of these pairwise arm differences reaches
significance under an exact paired McNemar test (single-pass vs per-dimension
$p{=}0.13$, per-dimension vs two-stage $p{=}0.22$, two-stage vs oracle
$p{=}0.13$; Appendix~\ref{app:ci}); the claim concerns the observed execution
path (every extra omission the oracle arm catches is a case the gated pipeline
had scored \texttt{unknown} at the gate), not a significant rate gap.

\paragraph{Cross-model robustness.}
Porting the unchanged pipeline across providers shows that high commission
detection is not model-specific (Appendix~\ref{app:matrix}): \emph{commission} detection is high
everywhere (gpt-4o $0.89$, Claude Opus~4.8 $1.00$, Gemini~2.5 Flash $0.91$,
Llama~3.1 8B $0.95$), but the clean-fixture flag rates differ substantially
(Table~\ref{tab:crossmodel}), so high detection alone does not show that each
model separates the flawed and clean versions reliably. What differs is the
operating point, and Table~\ref{tab:crossmodel_ops} shows where: Opus~4.8
rates the \emph{clean} fixture medium or high on 18 of 22 commission pairs
(false alarm $0.82$), the over-severity the human gold finds for gpt-4o, only
stronger; Llama~8B almost never abstains (one \texttt{unknown} in eleven
omissions) and calls the missing evidence high; Gemini is closest to gpt-4o
(26/33 identical injected verdicts). High commission detection is observed
with every tested model; the abstention behaviour and the false-alarm profile
are not, so changing the model requires a fresh assessment of calibration and
false alarms (\S\ref{sec:calib}).

Detection differs by flaw type (Table~\ref{tab:variants}). The
assessor catches $0.86$--$0.91$ of \emph{commission} flaws (flawed-but-present
evidence, the case adequacy reasoning targets) but only $0.45$ of \emph{omission}
flaws. The omission misses are not confident errors: the six missed variants all
abstain to \texttt{unknown} rather than assert low risk, because deleting the
supporting section makes retrieval fail. This is the same failure point we find on real papers
(\S\ref{sec:realpapers}): on these fixtures the assessor flags most
commission variants when the evidence is retrieved and abstains when it is
not. The only false-alarm signal ($0.09$ in every run) is one
clean-fixture dimension rated medium (\emph{inference} in run~1,
\emph{specification} in runs~2 and~3), the same over-severity we quantify
against the human gold (\S\ref{sec:vsgold}); the clean audit is itself not
deterministic. Table~\ref{tab:results} shows no false alarm on the original
eleven-flaw set.

\begin{table}[t]
  \centering\small
  \caption{The 33-variant benchmark (gpt-4o, mean of three runs at
  temperature~0; detection decisions identical on 32/33 variants). Adequacy reasoning
  catches \emph{commission} flaws; \emph{omission} misses are abstentions to
  \texttt{unknown}, marking the evidence-grounding bottleneck rather than wrong
  judgements. Intervals in Table~\ref{tab:ci}.}
  \label{tab:variants}
  \begin{tabular}{lccc}
    \toprule
    \rowcolor{tablehead} variant type & $n$ & detection & localization \\
    \midrule
    commission & 22 & 0.89 & 0.82 \\
    omission   & 11 & 0.45 & 0.33 \\
    \midrule
    \textbf{all} & 33 & \textbf{0.75} & \textbf{0.66} \\
    \bottomrule
  \end{tabular}
\end{table}

\subsection{Real papers: retrieval limits audit coverage}
\label{sec:realpapers}

We next audit the 26 papers tagged difference-in-differences in the
CausalVerify corpus~\cite{causalverify} (Appendix~\ref{app:corpus} gives its
composition and the limits of that tag). There is no
injected ground truth; we report
the raw per-dimension risk distribution of each assessor
(Figure~\ref{fig:realcorpus}). The keyword baseline is permissive: $91\%$ of its
judgements are \emph{low}. It sees topic vocabulary, declares the dimension
supported, and flags almost nothing. The two-stage LLM assigns a wider range of risk labels, and its
most informative output is that ${\sim}40\%$ of judgements are \texttt{unknown}:
the relevance gate could not surface evidence, so the system says so rather than
over-flagging. Two regimes appear: dimensions whose discussion is diffuse or
lives in figures the text retrieval cannot reach are dominated by
\texttt{unknown}, while dimensions where a section is retrieved are judged
inadequate. On real papers, then, retrieval limits coverage, and weakly
relevant passages are associated with high-risk ratings. These results
identify retrieval and the treatment of missing evidence as failure points; they
do not separate their contributions from rubric applicability and assessment
errors. The design therefore reports retrieval quality and an explicit
\texttt{unknown} state distinct from a substantive high-risk judgement.

\paragraph{Why fixtures abstain but real papers flag.}
The synthetic omission misses of \S\ref{sec:variants} become abstentions, yet
on real papers a missing check usually becomes a substantive \emph{high}. The
asymmetry is one branch of the gate. Retrieval is lexical
(Appendix~\ref{app:retrieval}): deleting a section from the short fixture leaves
no passage that even shares its vocabulary, the gate finds nothing relevant, and
the pipeline abstains without calling the judge. A full paper usually
contains a section that shares the vocabulary, so the gate returns
\emph{partial} relevance, the judge is called on an off-target passage, reports
the evidence as missing, and the prompt maps missing to high. The pilot
shows the pattern (Table~\ref{tab:gate}): all 22 failed-retrieval cells are
\texttt{unknown}, and 20 of the 22 weak-retrieval cells are \emph{high}, 19 of
them with \texttt{evidence\_status=missing}. This is the branch the calibration
layer of \S\ref{sec:calib} targets.

Judgements cite specific evidence
(Appendix~\ref{app:qual} gives a qualitative example). Roughly $44\%$ of
judgements on these papers are \emph{high}; without dimension-level
human labels beyond the five pilot papers we cannot say whether this is warranted or over-strict where
retrieval succeeds.

\subsection{An annotation pilot with reconciled labels}
\label{sec:gold}

To test agreement with human judgement, we built a small reconciled reference
set (the ``gold''): two annotators from outside the author team, who had no
part in building \sys{}, independently labelled five corpus papers $\times$ 11
dimensions (55 cells), without AI tools or sight of \sys{}'s output, and then reconciled the 13 cells on which they differed in
applicability or risk between themselves, without the first author
(Appendix~\ref{app:supp}). Their
independent labels agree on 45 of the 54
cells both rated (Cohen's $\kappa=0.56$; quadratic-weighted $\kappa=0.62$),
every disagreement is one severity step, and 23 of the 55 cells are labelled
under an \emph{analogue} reading of the dimension because the design is not a
canonical DID. The reconciled gold (55 cells) is low 7, medium 46, high 2; we
read this pilot as preliminary. The reviewed version
used labels two people had produced with LLM assistance and sent to the first
author, with no annotation record; they are replaced here (agreement with the
human gold 31/55, Appendix~\ref{app:supp}).

\subsection{\sys{} versus the human gold}
\label{sec:vsgold}

We use this gold as a \emph{diagnostic pilot}, not a validation of accuracy:
two high-risk gold cells cannot establish agreement with human judgement, but they localize
\emph{how} \sys{} fails
(Table~\ref{tab:vsgold}, Figure~\ref{fig:vsgold}). \sys{} abstains on 22/55
cells, and when it answers its errors are almost entirely one-directional:
25/33 answered cells are more severe than the reconciled gold, none less
severe; the gold contains two high-risk cells, whereas \sys{} assigns 24. With so few high
gold labels, precision and recall are noisy (of the two high gold cells \sys{}
flags one and abstains on the other). The
abstentions are retrieval failures, not absent evidence: both annotators
located reported evidence on 19 of the 22 abstained cells, at least one on 21.

\subsection{Calibration: a pre-specified rule, in-sample}
\label{sec:calib}

Of the re-run's 24 over-severe \emph{high} judgements, $23/24$ have
\texttt{evidence\_status=missing} and $20/24$ have
\texttt{retrieval\_quality=weak}. \sys{} treats ``weak retrieval found nothing''
as a substantive \emph{high}, conflating a \emph{retrieval failure} with the paper
lacking the evidence, which the reconciled human label puts at low or medium: a fixable
calibration target. The pattern re-emerges in every leave-one-paper-out fold
of the pilot (a recurrence check; no rule is refitted). (This uses a
re-run with retrieval-quality and evidence-status fields; it
differs from the corpus run of \S\ref{sec:vsgold}
in one cell; its pre-calibration figures are the same 25/33 over-severe and
exact agreement $0.24$.)

A deterministic rule over fields the LLM already produces reduces the
over-severity: a weak-retrieval \emph{high} (20 cells) is demoted to
\emph{medium} (keep coverage) or to \texttt{unknown} (abstain). Keeping
coverage, the rule raises exact agreement from $0.24$ to $0.76$ ($8/33$ to
$25/33$) and cuts over-severe cells from 25 to 8: on the 33 answered cells it
changes seventeen incorrect labels to the reconciled label and no correct
label to an incorrect one, and keeps the one high gold cell \sys{} had caught.
The abstention variant removes those 20 predictions (coverage $33/55 \to
13/55$) and leaves the remaining labels unchanged. Three further single-cell
rules lift exact agreement to $0.85$ without raising weighted $\kappa$ under
the demotion policy; they are post-hoc string and field rules, reported as
exploratory
(Appendix~\ref{app:supp}, Table~\ref{tab:calib},
Figure~\ref{fig:calib_tradeoff}). Exact agreement is a weak yardstick here: the gold is
$84\%$ \emph{medium}, and a constant \emph{medium} label would score $26/33$ on
the same cells; the statistic to read is weighted
$\kappa$, $0.08 \to 0.23$, with an interval that includes zero
(Appendix~\ref{app:ci}). The rules were fixed before the human labels arrived
but were derived on these same five papers against the earlier labels, so the
lift is in-sample: it points the same way in all five papers and is untested on
unseen papers.

\section{Discussion}
\label{sec:discussion}

\paragraph{What \sys{} measures.} A measurement is only as good as the link
between the construct and the task that operationalizes
it~\cite{jacobs2021measurement,raji2021benchmark}. \sys{} audits whether a
climate-policy study's identification assumptions are \emph{adequately supported
by its reported evidence}: reported-evidence adequacy is an operational proxy for
identification credibility, not causal validity itself.

\paragraph{Why climate policy.} Climate-policy estimates inform governance,
so a confidently wrong credibility judgement has a cost; \sys{} therefore
abstains when its gate finds no relevant passage.

\paragraph{Retrieval limits audit coverage.} A further gap is retrieving the
relevant evidence. On the
real-paper corpus it limits coverage, and in the pilot an assessor handed
weakly relevant evidence usually rates the dimension high (20 of 22
weak-retrieval pilot cells), so \sys{} reports retrieval quality and an
explicit \texttt{unknown} distinct from a substantive high-risk judgement.

\section{Conclusion}
\label{sec:conclusion}

The causal effect of a climate policy is unobservable; whether a study's
reported evidence supports its identification assumptions can be assessed.
\sys{} operationalizes that assessment with a bounded, evidence-grounded
pipeline, flaw injection as local ground truth, and a reconciled pilot whose human labels evaluate the auditor and score a
pre-specified rule that reduces its over-severity in-sample; its value
in expert review and on climate-policy papers is untested.

\section*{Limitations}
\label{sec:limitations}

Our human gold is pilot-scale: 5 papers and 55 cells from two annotators who
reconciled their differences themselves, so the comparison is a preliminary
signal rather than a calibrated benchmark; its labels are $84\%$ \emph{medium},
which makes exact agreement lenient and $\kappa$ the more informative statistic.
The gold is the reconciled judgement of two annotators, not of a panel of DID
methodologists; it has only two \emph{high} cells; 23 of its 55 cells judge an
analogue of the assumption because the design is not a canonical DID; every
agreement figure is conditional on the 33 cells \sys{} answered (it abstained
on 22, on 19 of which both annotators found evidence); and a constant
\emph{medium} label would out-score the calibrated system on exact agreement. The current scope is confined to
difference-in-differences identification; whether the rubric-plus-injection
methodology transfers to other quasi-experimental designs is untested. The rubric,
flaw taxonomy, and injector are co-designed, so synthetic detection rates are best
read as internal consistency under known threats rather than external
generalization (Appendix~\ref{app:flawlit} traces each injected flaw to the
literature that names it, so the threat catalogue is inherited while the
injector's prose is ours); flaws authored by independent experts are future
work. The benchmarks are small: at $n{=}11$ and $n{=}33$ no pairwise difference
between architecture arms reaches significance under a paired exact test, and
the human-gold intervals are wide (Appendix~\ref{app:ci}), so we report
descriptive differences on these benchmarks; the samples do not establish
population-level differences between the architecture variants. Those intervals and the
cell-level paired tests also treat the 55 pilot cells as independent although
they are nested in five papers, and the calibration rules were derived on the
pilot they are scored on, so the calibration lift is an in-sample result whose
size on unseen papers is untested. The parallel-trends criterion is likewise a
reporting check: \sys{} does not test whether a pre-trend test had the power to
detect a violation. The real corpus is a
general-economics stress test that contains no climate-policy evaluation in
the narrow sense (Appendix~\ref{app:corpus}); a climate-policy corpus run is
the natural next test. Relatedly,
flaw injection certifies the auditor against the \emph{injected} threat on a
single dimension only and inherits the taxonomy's blind spots, so we treat
detection and localization as necessary rather than sufficient, pairing them with
the false-alarm rate and the human-gold comparison. Evidence
grounding limits coverage. \sys{} emits \texttt{unknown} only when the relevance
gate finds no suitable passage; partial matches still reach the judge and can
produce over-severe verdicts. Finally, \sys{}
screens whether reported evidence supports a
study's identification assumptions and does not adjudicate the true causal effect,
which is never observed. Its outputs surface risk for expert review and do
not replace it.

\section*{Acknowledgments}

We thank Yan Li and Miao Zhang, the two annotators of the pilot in
\S\ref{sec:gold}, who read the five papers, labelled them independently and
reconciled their differences on their own time, and the anonymous reviewers,
whose comments prompted the uncertainty
quantification in Appendix~\ref{app:ci}, the account of the retrieval mechanism
in Appendix~\ref{app:retrieval}, and the analysis in \S\ref{sec:realpapers} of
why the system abstains on short fixtures yet flags missing evidence on full
papers. The icons in Figures~\ref{fig:overview} and~\ref{fig:arch} are from
\mbox{Flaticon.com}.

\appendix
\section{Domain Motivation and Design Rationale}
\label{app:motivation}

\subsection{Why climate-policy DID is a demanding audit setting}
Climate-policy evaluations can raise several identification challenges at
once. Policy announcement, legal adoption, and actual implementation may occur
at different dates, creating scope for anticipation. Carbon markets and
environmental regulations may be introduced at different times across
jurisdictions, so that heterogeneous and staggered treatment effects need to
be handled. Emissions, investment, and production responses can spill across
regional borders, challenging no-interference assumptions. Climate policies
can overlap with industrial, energy, and air-pollution regulations,
complicating the exclusion of concurrent interventions. Policy effects may
emerge with long and heterogeneous lags. Emissions and environmental outcomes
may be assembled from administrative, modelled, or remotely sensed sources whose
measurement properties vary across places and time. These features do not make
DID unique to climate policy; they motivate a rubric that makes the evidence
for each design choice explicit.

\subsection{Identification auditing and research reliability}
\sys{} addresses one component of a broader research-reliability problem: published
conclusions can vary with specification choices, undisclosed analytic flexibility,
and researcher degrees of freedom
\cite{leamer1983,ioannidis2005,simmons2011,gelmanloken2014,silberzahn2018many,brodeur2020methods,botviniknezer2020variability}.
This is the identification-credibility component of the broader research-design or
``credibility revolution'' program in applied econometrics
\cite{angristpischke2010,atheyimbens2017}, which increasingly treats an assumption
like parallel trends as a matter of degree rather than a binary
\cite{rambachanroth2023,roth2022}. \sys{} does not attempt to audit every component
of an empirical paper; it focuses on the assumptions that support a causal
interpretation.

\subsection{Why the pipeline is bounded}
Open-ended agents may choose different tools and reasoning paths across runs, making
their outputs difficult to reproduce and compare. Large language models are
increasingly used for evaluative and reviewing tasks
\cite{zheng2023judge,liang2024feedback,lu2024aiscientist}, but as open-ended agents
their reasoning traces are hard to audit. \sys{} instead fixes the pipeline order in
code and confines model-driven decisions to evidence retrieval and adequacy
assessment. Each model call operates under a fixed tool set, a hard step budget, and
a logged trace. This design trades autonomy for inspectability and controlled
evaluation under flaw injection; it fixes the sequence of operations, not the
model's outputs.

\section{Flaw Taxonomy Provenance}
\label{app:flawlit}
The rubric, taxonomy, and injector are co-designed, but the catalogue of
threats is not ours: each injected flaw is a threat named in the DID and
research-design literature (Table~\ref{tab:flawlit}). What we authored is the
injector's prose for each threat, which is why we describe synthetic detection
as internal consistency under known threats and treat independently authored
flaws as future work.

\begin{table}[t]
  \centering\small
  \caption{Each injected flaw, its target dimension, and the literature that
  names the threat.}
  \label{tab:flawlit}
  \setlength{\tabcolsep}{4pt}
  \resizebox{\columnwidth}{!}{%
  \begin{tabular}{llp{4.2cm}}
    \toprule
    \rowcolor{tablehead} flaw & dimension & threat named in \\
    \midrule
    pretrend divergence & parallel trends & \cite{roth2022,roth2023,rambachanroth2023} \\
    anticipation effect & no anticipation & \cite{callaway2021,roth2023} \\
    forbidden comparison & staggered timing & \cite{goodmanbacon2021,dechaisemartin2020,sunabraham2021} \\
    spillover contamination & SUTVA / spillovers & \cite{atheyimbens2017,roth2023} \\
    noncomparable controls & control group & \cite{callaway2021,atheyimbens2017} \\
    bad control & specification & \cite{angristpischke2009} \\
    understated standard errors & inference & \cite{bertrand2004trust} \\
    cherry-picked window & sample period & \cite{simmons2011,brodeur2020methods} \\
    confounding policy & concurrent policies & \cite{meyer1995} \\
    missing placebo & placebo / robustness & \cite{bertrand2004trust,atheyimbens2017} \\
    measurement break & data measurement & \cite{meyer1995} \\
    \bottomrule
  \end{tabular}}
\end{table}

\section{Ablation Matrix and Cross-Model Panel}
\label{app:matrix}

Table~\ref{tab:matrix} grades compute on the 33-variant benchmark:
per-dimension prompting (one adequacy call per dimension) detects more variants
than a single joint call, and the retrieval-and-gate layer converts omission
detection into abstention.
Table~\ref{tab:crossmodel} ports the unchanged pipeline across providers:
commission detection is uniformly high, while false-alarm and localization
profiles vary sharply by model. Cross-provider decoding defaults differ (newer
reasoning models reject an explicit temperature), so these are single runs.
Table~\ref{tab:crossmodel_ops} shows where the operating points diverge: how
each model treats a deleted section, how often it already rates the unmodified
fixture medium or high, and how often its injected-variant verdict matches
gpt-4o's.

On real papers the gate's behavioural contrast sharpens. This run used a second
single-pass implementation, distinct from the strict-schema arm of
Tables~\ref{tab:results} and~\ref{tab:matrix}: its prompt permits
\texttt{unknown} with the instruction to ``use sparingly''
(Appendix~\ref{app:prompts}). Over the 25 papers common to both runs ($275$
cells) it chose \texttt{unknown} once ($0.4\%$; a model-written abstention with
a rationale, not a harness default), where the gated pipeline abstains
mechanically on $38.9\%$. The gated pipeline's \texttt{unknown} cells become
low ($1.8\%\!\to\!14.9\%$) or medium ($14.2\%\!\to\!43.6\%$) judgements while
the high-risk share stays comparable (gated $45.1\%$, single pass $41.1\%$). These shares
describe behaviour (a discretionary abstention option is rarely taken, a
mechanical one often is); without cell-level gold they do not show which
judgements are more reliable.

\begin{table}[t]
  \centering\small
  \caption{Compute-graded ablation on the 33-variant benchmark (gpt-4o).
  det / fa / loc = detection, false alarm, localization. Calls per paper are
  the nominal maximum: when the gate finds nothing relevant the adequacy call
  is skipped, and retries are not counted. Localization takes the top-ranked
  flagged dimension; ties in risk level are broken by the fixed rubric order.}
  \label{tab:matrix}
  \setlength{\tabcolsep}{4pt}
  \resizebox{\columnwidth}{!}{%
  \begin{tabular}{lcccc}
    \toprule
    \rowcolor{tablehead} arm & calls/paper & det & fa & loc \\
    \midrule
    full-paper single pass & 1 & 0.727 & 0.000 & 0.727 \\
    per-dimension, no retrieval & 11 & 0.879 & 0.000 & 0.879 \\
    two-stage (retrieval + gate) & 22 & 0.747 & 0.091 & 0.657 \\
    \bottomrule
  \end{tabular}}
\end{table}

\begin{table}[t]
  \centering\small
  \caption{Cross-model panel on the 33-variant benchmark (single run per
  model; gpt-4o is the three-run mean; single runs support the qualitative
  porting claim only). comm/omit = commission/omission detection. Intervals in
  Table~\ref{tab:ci}.}
  \label{tab:crossmodel}
  \setlength{\tabcolsep}{3.5pt}
  \resizebox{\columnwidth}{!}{%
  \begin{tabular}{lccccc}
    \toprule
    \rowcolor{tablehead} model & det & fa & loc & comm & omit \\
    \midrule
    gpt-4o & 0.75 & 0.09 & 0.66 & 0.89 & 0.45 \\
    Claude Opus 4.8 & 0.91 & 0.82 & 0.24 & 1.00 & 0.73 \\
    Gemini 2.5 Flash & 0.79 & 0.18 & 0.36 & 0.91 & 0.55 \\
    Llama 3.1 8B (local) & 0.94 & 0.27 & 0.27 & 0.95 & 0.91 \\
    \bottomrule
  \end{tabular}}
\end{table}

\begin{table}[t]
  \centering\small
  \caption{Cross-model operating points on the 33-variant benchmark (single
  runs; gpt-4o run 1). Omission column: verdicts on the eleven deleted-section
  variants, in the order \texttt{unknown} / high / medium / low. Clean column:
  commission pairs (of 22) whose \emph{unmodified} fixture is already rated
  medium or high. Last column: exact agreement with gpt-4o's injected-variant
  verdict.}
  \label{tab:crossmodel_ops}
  \setlength{\tabcolsep}{3.5pt}
  \resizebox{\columnwidth}{!}{%
  \begin{tabular}{lccc}
    \toprule
    \rowcolor{tablehead} model & omission verdicts & clean $\geq$ med & agree \\
    \midrule
    gpt-4o & 6 / 3 / 2 / 0 & 2 / 22 & -- \\
    Claude Opus 4.8 & 3 / 7 / 1 / 0 & 18 / 22 & 20 / 33 \\
    Gemini 2.5 Flash & 4 / 3 / 3 / 1 & 4 / 22 & 26 / 33 \\
    Llama 3.1 8B (local) & 1 / 10 / 0 / 0 & 6 / 22 & 19 / 33 \\
    \bottomrule
  \end{tabular}}
\end{table}

\section{Retrieval Details}
\label{app:retrieval}
The only corpus queried is the paper under audit; there is no external corpus
and no embedding index. Papers are ingested through a light markdown convention
(\texttt{\#\#} section headings plus \texttt{@figure}/\texttt{@table} caption
lines), and the searchable blocks are the title, abstract, each section, and
each figure or table caption. On the LLM path retrieval is section-level and
lexical: each dimension has a fixed query pack of 11--19 distinct terms (e.g.\
\emph{pre-trend, event study, lead coefficients} for parallel trends), a block
scores by the summed frequency of its matched terms, and the top $k{=}4$ blocks
are kept, each truncated to 2,000 characters. The relevance gate sees the first
1,200 characters of each candidate and labels it \emph{high}, \emph{partial},
or \emph{none}; retrieval quality is \emph{good} if any candidate is high,
\emph{weak} if only partial candidates remain, and \emph{failed} otherwise. A
failed dimension is scored \texttt{unknown} without an adequacy call; weak and
good dimensions pass the kept blocks, in full, to the adequacy judge together
with the rubric's assumption, implication, and expected evidence. The keyword
baseline instead searches paragraph chunks capped at 900 characters
(sentence-split when longer) with a single query string and top $k{=}5$, scoring
positive and negative sentinel phrases. With \texttt{max\_steps}{=}1 there is
no iterative exploration, so breadth is the top-$k$ cap and depth is one round.
Chunk size, $k$, and embedding retrieval were not swept. The oracle-retrieval
arm of \S\ref{sec:variants} is a diagnostic comparison in which the target
section is supplied directly and the gate is bypassed; it is not a formal upper
bound on retrieval performance.
Table~\ref{tab:gate} shows how the gate's three outcomes map to risk on the
pilot-gold cells, the mechanism behind the fixture/real-paper asymmetry of
\S\ref{sec:realpapers}.

\begin{table}[t]
  \centering\small
  \caption{Retrieval quality versus emitted risk on the 55 pilot-gold cells
  (re-run recording retrieval quality and evidence status). Failed retrieval always abstains; weak retrieval almost always
  becomes a substantive \emph{high}, 19 of those 20 with
  \texttt{evidence\_status=missing}.}
  \label{tab:gate}
  \begin{tabular}{lcccc}
    \toprule
    \rowcolor{tablehead} retrieval quality & \texttt{unknown} & high & medium & low \\
    \midrule
    failed (22) & 22 & 0 & 0 & 0 \\
    weak (22)   & 0 & 20 & 2 & 0 \\
    good (11)   & 0 & 5 & 5 & 1 \\
    \bottomrule
  \end{tabular}
\end{table}

\section{Prompts}
\label{app:prompts}
All gpt-4o calls run at temperature~$0$ (other providers:
Appendix~\ref{app:repro}); all but the real-paper single-pass
variant at the end of this appendix use a strict JSON schema. Angle-bracketed
fields are substituted from the rubric entry or the retrieved passages; nothing
else varies across dimensions or papers. The per-dimension no-retrieval arm of
\S\ref{sec:variants} reuses the adequacy prompt below verbatim, passing the whole
paper as the evidence items instead of retrieved sections, so that arm differs
from the full pipeline only in what the prompt is fed.

\promptlabel{Relevance gate (system).}
\begin{promptbox}
You are screening retrieved passages for a difference-in-differences (DID)
identification audit. For ONE identification dimension, decide whether each
passage is actually evidence about THAT dimension's assumption. This is a
relevance check, NOT a quality or risk judgement. Mark a passage 'high' if it
directly concerns the assumption (e.g. an event-study / pre-trend discussion for
parallel trends), 'partial' if it touches it indirectly, and 'none' if it is
about something else. Be strict: a passage that merely shares vocabulary but is
about a different topic is 'none'.
\end{promptbox}

\promptlabel{Relevance gate (user).}
\begin{promptbox}
Identification dimension: <name>\\
Assumption: <assumption>\\
Testable implication: <implication>\\[5pt]
Candidate passages:\\
{}[0] (<source>) <passage text, first 1200 characters>\\
{}[1] (<source>) <...>\\[5pt]
Classify each passage's relevance to this dimension.
\end{promptbox}
\begin{sloppypar}\noindent Schema: one \texttt{\{index, relevance, reason\}}
object per passage, with \texttt{relevance} in
\texttt{\{high, partial, none\}}.\end{sloppypar}

\promptlabel{Adequacy judge (system).}
\begin{promptbox}
You audit the causal-IDENTIFICATION credibility of difference-in-differences
(DID) studies in environmental policy evaluation. You do NOT judge whether the
estimated effect is true \textemdash{} the counterfactual is never observed. For
ONE identification dimension, judge whether the evidence the paper reports is
ADEQUATE to support the assumption.\\[5pt]
Crucial distinction: evidence PRESENCE is not evidence ADEQUACY. The mere
mention of a topic (an event study, a balance table, 'placebo') does not make the
assumption supported \textemdash{} assess whether the reported evidence actually
establishes the testable implication, and whether it reveals a flaw.\\[5pt]
Return a risk level and an evidence-status label:\\
\hspace*{1em}risk: low | medium | high\\
\hspace*{1em}evidence\_status: sufficient | partial | missing | flawed\\
Map roughly: sufficient->low, partial->medium, missing->high, flawed->high. Be
conservative: if the evidence is absent, thin, or describes a flawed design, do
not call it supported. Cite the specific evidence (by source) your judgement
rests on. Keep the rationale to one or two sentences.
\end{promptbox}

\promptlabel{Adequacy judge (user).}
\begin{promptbox}
Identification dimension: <name>\\
Assumption: <assumption>\\
Testable implication: <implication>\\
Evidence a credible paper would report: <expected evidence>\\[5pt]
Evidence retrieved from THIS paper:\\
- [<source>] <passage text>\\[5pt]
Judge the adequacy of this evidence for the assumption.
\end{promptbox}
\noindent When the gate keeps nothing the evidence block reads \texttt{(no
evidence was retrieved for this dimension)}, but that path is unreachable in the
deployed pipeline: a failed gate short-circuits to \texttt{unknown} without a
model call (\S\ref{sec:realpapers}).
\begin{sloppypar}\noindent Schema:
\texttt{\{risk, evidence\_status, rationale, cited\_evidence\}}.\end{sloppypar}

\promptlabel{Single-pass ablation, synthetic benchmarks (system and user).}
\begin{promptbox}
You audit the causal-identification credibility of a difference-in-differences
(DID) study. You do NOT judge whether the estimated effect is true. For EACH of
the eleven identification dimensions listed, judge whether the evidence the paper
reports is ADEQUATE to support that assumption. Evidence presence is not evidence
adequacy. Return one judgement per dimension.
\end{promptbox}
\begin{promptbox}
Identification dimensions to judge:\\
- <dimension id>: <assumption> (expect: <expected evidence>)\\
\hspace*{1em}[... eleven lines ...]\\[5pt]
THE PAPER:\\
{}[<source>]\\
<full section text>\\[5pt]
Judge the adequacy of the reported evidence for EVERY dimension.
\end{promptbox}
\begin{sloppypar}\noindent Schema: a \texttt{judgements} array of
\texttt{\{dimension, risk, evidence\_status, rationale\}}, with
\texttt{dimension} restricted to the eleven rubric ids and \texttt{risk} to
\texttt{low | medium | high}; this arm cannot answer \texttt{unknown}. It
produced the single-pass rows of Tables~\ref{tab:results}
and~\ref{tab:matrix}.\end{sloppypar}

\promptlabel{Single-pass variant, real-paper run (system, abridged to the risk block).}
\begin{promptbox}
For each dimension output one risk level:\\
\hspace*{1em}low : adequate, credible evidence is reported.\\
\hspace*{1em}medium : evidence is partial, indirect, or weak.\\
\hspace*{1em}high : the required evidence is absent, or present but flawed.\\
\hspace*{1em}unknown : the text gives no basis to judge either way (use sparingly).\\[5pt]
Also output evidence\_status in \{sufficient, partial, missing, flawed\}
(roughly sufficient->low, partial->medium, missing/flawed->high). Be
conservative: absent, thin, or flawed evidence is not 'supported'.
\end{promptbox}
\begin{sloppypar}\noindent The real-paper contrast of
Appendix~\ref{app:matrix} used this second implementation
(\texttt{single\_pass\_assessor.py}). Its preamble restates the adequacy
standard of the judge prompt above, it receives the whole paper with the
rubric, and it returns free JSON keyed by dimension id in which a missing or
malformed cell is coerced to \texttt{unknown}.\end{sloppypar}

\section{Reproducibility Details}
\label{app:repro}
\begin{sloppypar}\noindent Code, configuration, prompts, committed run files, the
frozen pilot inputs, the annotation protocol and the locked gold are available
at \url{https://github.com/yonghongzhang-io/ARGUS}.\end{sloppypar}

All gpt-4o stages use temperature~$0$ with \texttt{max\_steps}{=}1; newer
reasoning models (Claude Opus~4.8) reject an explicit temperature and run at
provider defaults. The identification rubric, flaw taxonomy, and
33-variant catalogue are declarative configuration files in the code base, and
the keyword baseline plus all leak-freedom (non-circularity) checks are
deterministic and run without model access. Eleven-flaw numbers are July-2026
runs of the current runner; detection replicates exactly across the
\texttt{gpt-4o} alias and the dated \texttt{gpt-4o-2024-11-20} snapshot
(localization varies between runs, $0.455$--$0.727$, and the snapshot run
rates one clean dimension medium, a false alarm of $0.09$ where the reported
run has none). A June-2026 run of the
same code had scored two-stage $1.000$/$0.909$ and single-pass $0.455$; neither
value was reproduced in July on the alias \emph{or} the dated snapshot. The
available records do not establish the cause of this difference, so we report
the later runs with their dates and model identifiers. The 33-variant benchmark was run three
times end-to-end (roughly $750$ model calls per run, ${\sim}20$ minutes and
${\sim}$US\$8 per gpt-4o run; the local Llama run took ${\sim}67$ minutes on a
laptop at zero API cost); binary detection agreed on 32/33 variants across the
three runs and target-dimension risk labels on 31/33, while the clean audit
flagged a different dimension in run~1 than in runs~2 and~3, and Table~\ref{tab:variants} reports the three-run mean. The
two-call-per-dimension pipeline audits one paper for roughly US\$0.25 of
gpt-4o usage.

\section{Uncertainty Quantification}
\label{app:ci}
Table~\ref{tab:ci} gives Wilson 95\% intervals for the main proportions reported in
the main text and exact tests between arms evaluated on the same items;
Table~\ref{tab:ci_gold} gives cell-bootstrap intervals (5,000 resamples) for
the human-gold agreement metrics. At $n{=}11$ the keyword-versus-two-stage gap
is marginal under the paired test on the same flaws (7 vs 1 discordant,
$p{=}0.07$; an unpaired Fisher test, which ignores the pairing, gives
$p{=}0.03$); at $n{=}33$ no
pairwise difference between the single-pass, per-dimension, two-stage, and
oracle arms reaches significance. For the calibration lift the comparison is paired, so overlap of the
intervals in Table~\ref{tab:ci_gold} is not the test. On the 33 answered cells
the weak-retrieval rule alone turns seventeen wrong cells exact and none the
other way (exact McNemar $p{<}0.001$); all four rules turn twenty
($p{<}0.001$). Those tests treat cells as independent, yet the cells are nested
in five papers. By paper, the single rule improves exact agreement in all five
papers, as do the four rules (sign test $p{=}0.0625$ in each case, the smallest
value five papers allow). Neither analysis removes the bias of deriving the
rules on the pilot they are scored on, and both count agreement with a gold
that is $84\%$ \emph{medium}, which a rule that demotes to \emph{medium} matches
by construction. We therefore claim a consistent in-sample direction, not a
demonstrated out-of-sample gain.

\begin{sloppypar}\noindent Tables~\ref{tab:ci} and~\ref{tab:ci_gold} are
computed from committed files by \path{uncertainty.py} and
\path{calibration_recheck.py} in \path{experiments/ablations/}; the pilot cells
they read are frozen with checksums in
\path{experiments/annotation/pilot_frozen/} and the locked gold in
\path{experiments/annotation/gold_final/}.\end{sloppypar}

\begin{table}[t]
  \centering\small
  \caption{Wilson 95\% confidence intervals (CI) for reported rates (top) and exact tests
  between arms (bottom; McNemar tests are paired on the same flaws or
  variants; discordant = detected by the first arm only / by the second only). False-alarm rows give counts only: every variant is paired with
  the same clean fixture, so they are repeated readings of one clean audit of
  eleven dimensions, not independent trials, and an interval would overstate
  what they show. Detection intervals describe this fixed test set (33 variants
  nested in eleven dimensions of one fixture), not real papers at large.}
  \label{tab:ci}
  \setlength{\tabcolsep}{4pt}
  \resizebox{\columnwidth}{!}{%
  \begin{tabular}{llcc}
    \toprule
    \rowcolor{tablehead} benchmark & quantity & $k/n$ & 95\% CI \\
    \midrule
    11 flaws & keyword detection & 2/11 & 0.05--0.48 \\
             & single-pass detection & 9/11 & 0.52--0.95 \\
             & per-dimension detection & 11/11 & 0.74--1.00 \\
             & two-stage detection & 8/11 & 0.43--0.90 \\
             & two-stage false alarm & 0/11 & -- \\
    33 variants & single-pass detection & 24/33 & 0.56--0.85 \\
             & per-dimension detection & 29/33 & 0.73--0.95 \\
             & two-stage detection (run 1) & 25/33 & 0.59--0.87 \\
             & \quad commission & 20/22 & 0.72--0.97 \\
             & \quad omission & 5/11 & 0.21--0.72 \\
             & two-stage false alarm & 3/33 & -- \\
             & oracle-retrieval detection & 30/33 & 0.76--0.97 \\
    cross-model & Opus 4.8 detection & 30/33 & 0.76--0.97 \\
             & Opus 4.8 false alarm & 27/33 & -- \\
             & Gemini 2.5 Flash detection & 26/33 & 0.62--0.89 \\
             & Gemini 2.5 Flash false alarm & 6/33 & -- \\
             & Llama 3.1 8B detection & 31/33 & 0.80--0.98 \\
             & Llama 3.1 8B false alarm & 9/33 & -- \\
    \midrule
    \rowcolor{tablehead} test & arms & discordant & $p$ \\
    \midrule
    Fisher & keyword vs two-stage (11) & -- & 0.030 \\
    McNemar & keyword vs two-stage (11) & 1 / 7 & 0.070 \\
    McNemar & single-pass vs per-dim.\ (33) & 1 / 6 & 0.125 \\
    McNemar & per-dim.\ vs two-stage (33) & 5 / 1 & 0.219 \\
    McNemar & single-pass vs two-stage (33) & 5 / 6 & 1.000 \\
    McNemar & two-stage vs oracle (33) & 1 / 6 & 0.125 \\
    \bottomrule
  \end{tabular}}
\end{table}

\begin{table}[t]
  \centering\small
  \caption{Cell-bootstrap 95\% intervals for agreement with the reconciled
  gold (answered cells only; quadratic-weighted $\kappa$; 5,000 resamples).
  Rule~1 is the weak-retrieval rule; rules~1--4 add the three single-cell rules
  of Appendix~\ref{app:supp}. Resampling cells ignores their nesting in five
  papers and may understate uncertainty.}
  \label{tab:ci_gold}
  \setlength{\tabcolsep}{4pt}
  \resizebox{\columnwidth}{!}{%
  \begin{tabular}{lccc}
    \toprule
    \rowcolor{tablehead} policy & $n$ & exact agreement & weighted $\kappa$ \\
    \midrule
    before & 33 & 0.24 (0.09--0.39) & 0.08 ($-0.05$--0.25) \\
    rule 1, demote$\to$med & 33 & 0.76 (0.61--0.88) & 0.23 ($-0.22$--0.63) \\
    rule 1, abstain$\to$unk & 13 & 0.62 (0.38--0.85) & 0.29 ($-0.18$--0.77) \\
    rules 1--4, demote$\to$med & 33 & 0.85 (0.73--0.97) & 0.22 ($-0.42$--0.79) \\
    rules 1--4, abstain$\to$unk & 13 & 0.85 (0.62--1.00) & 0.32 ($-0.52$--1.00) \\
    \bottomrule
  \end{tabular}}
\end{table}

\section{The Real-Paper Corpus}
\label{app:corpus}

The 26 papers audited in \S\ref{sec:realpapers} are those tagged
difference-in-differences in the corpus of 259 economics papers released with
CausalVerify~\cite{causalverify} (Table~\ref{tab:corpus}; titles in
Table~\ref{tab:corpuslist}). The tag is the corpus's method-family label, which
we did not verify paper by paper, and it is imperfect: one entry is a literature
review (below) and one identifies its effect from sibling comparisons. Nineteen entries are journal articles, one of them in a
medical journal, and seven are held as National Bureau of Economic Research
(NBER) working-paper versions. That corpus as released contains one
byte-identical duplicate (a 2002 Clean Air Act study under two identifiers);
both copies received identical
audits on all eleven dimensions from both assessors, and we count the paper
once, so every count reads 26 papers and 286 cells. One further entry is a
review of the school-spending literature rather than a primary DID study; we
retain and note it. By topic, the corpus spans finance and banking (5), health
(4), education (4), environment and energy (4), international trade (3), labour
(3), crime and social policy (2), and industrial location (1); the
environment-and-energy subset covers the industrial-activity effects of
environmental regulation, air quality and early-life mortality, nonlinear
electricity pricing, and gasoline prices and vehicle choice. None of the 26 is
a climate-policy evaluation in the narrow sense (a carbon price, an emissions
trading scheme, or an emission standard): the real corpus is a
general-economics stress test of the audit architecture, and the
climate-specific content of the paper is the rubric instantiation of
\S\ref{sec:framework} and Appendix~\ref{app:motivation}; the synthetic fixture
of \S\ref{sec:evaluation} is a generic environmental-policy study (a China
emissions-trading fixture ships with the code but produced none of the numbers
reported here). The environment-and-energy subset has an \texttt{unknown}
share of $0.43$ (44 cells), against $0.39$ in the remaining papers (242
cells); the corresponding high-risk shares are $0.34$ and $0.46$. With four
papers in the subset, this comparison does not establish whether audit
behaviour differs by domain.

\begin{table}[t]
  \centering\small
  \caption{Venue of the 26 audited papers, as recorded for the version held in
  the source corpus. Two entries carry a corrupted venue field in the corpus
  index; they are counted under the journal that published them (Journal of
  Political Economy, 2003; Review of Economic Studies, 2002). AEJ: American
  Economic Journal; JF: Journal of Finance; REStat: Review of Economics and
  Statistics; REStud: Review of Economic Studies.}
  \label{tab:corpus}
  \setlength{\tabcolsep}{4pt}
  \resizebox{\columnwidth}{!}{%
  \begin{tabular}{lc}
    \toprule
    \rowcolor{tablehead} venue / version & papers \\
    \midrule
    American Economic Review & 4 \\
    Quarterly Journal of Economics & 3 \\
    Journal of Political Economy & 2 \\
    Journal of Financial Economics & 2 \\
    AEJ: Applied Economics & 2 \\
    JF / REStat / REStud / J.\ Monetary Econ.\ / J.\ Int.\ Econ.\ (one each) & 5 \\
    New England Journal of Medicine & 1 \\
    NBER working-paper versions & 7 \\
    \bottomrule
  \end{tabular}}
\end{table}

\begin{table}[t]
  \centering\scriptsize
  \caption{The 26 audited studies, by title as indexed in the source corpus
  (the duplicated entry is listed once).}
  \label{tab:corpuslist}
  \setlength{\tabcolsep}{3pt}
  \begin{tabular}{p{0.94\columnwidth}}
    \toprule
    The Effect of the Banking Panic on the Supply of Credit to the Corporate Sector \\
    The Effects of State-Level Banking Competition on Innovation \\
    Do Credit Market Shocks Affect the Real Economy? Quasi-Experimental Evidence from the Great Recession and `Normal' Economic Times \\
    Cyclicality of Credit Supply: Firm Level Evidence \\
    Does Stock Liquidity Enhance or Impede Firm Innovation? \\
    Minimum Wages and Employment: A Case Study of the Fast Food Industry in New Jersey and Pennsylvania \\
    Minimum Wage Effects Across State Borders: Estimates Using Contiguous Counties \\
    Hospital Ownership and Public Medical Spending \\
    The Impact of Public Disclosure of Health Outcomes on Patient and Social Welfare \\
    Did Community Rating Induce an Adverse Selection Death Spiral? Evidence from New York, Pennsylvania, and Connecticut \\
    Mortality and Access to Care among Adults after State Medicaid Expansions \\
    Early Childhood Intervention and Life-Cycle Skill Development: Evidence from Head Start \\
    The Impacts of Environmental Regulations on Industrial Activity: Evidence from the 1970 and 1977 Clean Air Act Amendments and the Census of Manufactures \\
    Legalized Abortion and Crime \\
    Do Consumers Respond to Marginal or Average Price? Evidence from Nonlinear Electricity Pricing \\
    Estimates of the Impact of Crime Risk on Property Values from Megan's Laws \\
    Effects of Liberalized Trade on Plant Productivity \\
    The Surprisingly Swift Decline of U.S. Manufacturing Employment \\
    Factor Immobility and Trade: Poverty in India \\
    Trade Liberalization and Industry Wage Premia: Evidence from Colombia \\
    The Effect of State Policies on the Location of Manufacturing: Evidence from State Borders \\
    Air Quality and Early-Life Mortality: Evidence from Indonesia's Wildfires \\
    Does School Spending Matter? The New Literature on an Old Question \\
    The Evolution of Black/White Earnings: 1960--1980 \\
    The Effects of Class Size on Student Achievement: New Evidence from Population Variation \\
    Are Consumers Myopic? Evidence from New and Used Car Purchases \\
    \bottomrule
  \end{tabular}
\end{table}

\section{Qualitative Audit Examples}
\label{app:qual}

\paragraph{A complete audit report.}
Table~\ref{tab:casestudy} shows \sys{}'s actual output on the synthetic
fixture with the \emph{anticipation} flaw injected (\S\ref{sec:evaluation}): the
report flags exactly the damaged dimension as high, giving the anticipation-gap
rationale, keeps nine supported dimensions at low, and shows a mild
over-severity trace on \emph{inference} (the fixture does not state the number
of clusters), the same bias the human-gold pilot quantifies. This is the
dimension-level, evidence-cited report a reviewer receives.

\begin{table}[t]
  \centering\small
  \caption{\sys{} audit report (abridged) on the fixture with the injected
  anticipation flaw. Rationales are abridged from the evidence-cited report; the injected
  dimension is the one flagged high.}
  \label{tab:casestudy}
  \setlength{\tabcolsep}{4pt}
  \resizebox{\columnwidth}{!}{%
  \begin{tabular}{llp{5.2cm}}
    \toprule
    \rowcolor{tablehead} dimension & risk & rationale (abridged) \\
    \midrule
    no anticipation & \textbf{high} & outcomes began changing between the
    policy announcement and its rollout, undermining no-anticipation \\
    inference & medium & clustering and wild-cluster bootstrap reported, but the
    number of clusters is not stated \\
    parallel trends & low & event-study with null pre-period coefficients and
    placebo tests \\
    staggered timing & low & heterogeneity-robust estimators with a
    Goodman-Bacon decomposition \\
    \emph{7 further dimensions} & low & supporting evidence retrieved and judged
    adequate \\
    \bottomrule
  \end{tabular}}
\end{table}

\paragraph{A grounded judgement on a real paper.}
On a trade-and-wages paper \sys{} retrieves the employment-results section and
flags \emph{inference} as high because ``the paper only mentions
heteroskedastic-consistent standard errors and does not provide information on
clustering or the number of clusters'': a specific econometric observation, not
a keyword hit. Elsewhere it distinguishes a \emph{related but different} reported
check (an alternative-specification section) from the placebo evidence the
rubric specifies for that dimension.

\section{Supplementary Figures and Tables}
\label{app:supp}

\paragraph{Pilot-gold construction (\S\ref{sec:gold}).}
The protocol and analysis plan were written and committed before any label was
returned (\path{annotation/human_pilot/PROTOCOL.md}); each returned workbook
is recorded by SHA-256 on receipt, and the gold is locked once
(\path{experiments/annotation/gold_final/}). The two annotators, a student at
the first author's university and a member of a vocational college in China,
neither an author nor involved in building \sys{}, read the five PDFs
themselves, used no AI tool, saw no \sys{} output, did not confer before
returning, and signed to that effect in the workbook. The five papers are \texttt{paper\_01}, \texttt{03}, \texttt{07}, \texttt{08} and
\texttt{10} of the repository's corpus manifest, chosen in June 2026 before any
label existed. Independent labels: 55
cells each; applicability agrees on 49, with one cell rated not applicable by
one annotator; of the 54 cells both rated, 45 agree exactly (Cohen's
$\kappa=0.56$; quadratic-weighted $\kappa=0.62$), and all nine risk
disagreements are a single severity step apart. The 13 cells that differed in
applicability or risk were settled by the two annotators together, all 13 as
agreed decisions, 9 of them under an analogue reading of the dimension; no
third person was needed. Each annotator's verbatim quotes were checked against
the PDFs: 50/55 and 49/55 occur verbatim, the rest as partial matches, one cell
each not found. Three deviations are logged: one annotator first returned a
paper sheet pasted from another paper and corrected it herself within sixteen
minutes; the file first recorded for her had been re-saved on the first
author's computer (how is not known) and differed from her own last version in
eleven evidence-location cells and one rationale, in no label, quote or
confidence, so the record was rebuilt from her own file and none of the twelve
cells enters the analysis; and the other annotator first returned the workbook
without the sign-off sheet and re-sent it signed.

\paragraph{Change from the reviewed version.}
The reviewed version described two annotators and an adjudication round. No
record of how those labels were produced survives: the first author's
recollection is that two people produced them with LLM assistance and sent them
to the first author, who entered them; there is no file, sign-off or
adjudication record of theirs, so the reviewed description was not supported by
any record, and the labels are not treated as an annotation. They are kept,
checksummed, in \path{experiments/annotation/pilot_frozen/} and are replaced
here. Against the human gold they agree on 31/55 cells (Cohen's $\kappa=0.23$),
being less severe on 23 (low where the gold is medium); under them \sys{} was
more severe on 28/33 answered cells and exact on 5, against 25/33 and 8 now, and
the reviewed calibration figures (exact agreement $0.12 \to 0.36$, $0.45$ with
four rules) are superseded by \S\ref{sec:calib}: the larger lift now reflects
the medium-heavy gold, not the rule.

In the leave-one-paper-out check (\S\ref{sec:calib}), the demotion rule's
trigger conditions recur in all five folds: among each fold's over-severe
\emph{high} cells, $79$--$86\%$ have weak retrieval and $94$--$100\%$ have
missing evidence (computed on the rich re-run of \S\ref{sec:calib}). The folds test whether the error pattern recurs; no rule was refitted
on four papers and scored on the fifth.

\paragraph{The calibration rules.}
The layer acts on \emph{high} cells only and never raises a risk; the first
matching rule applies. Rule~1 (20 cells): \texttt{retrieval\_quality=weak}
$\to$ \emph{medium}, or $\to$ \texttt{unknown} under the abstain policy. Three
further rules each fire on one cell and demote to \emph{medium}: rule~2, the
dimension is staggered timing and the rationale names Callaway, Sun--Abraham,
Goodman-Bacon, or ``modern estimator''; rule~3, the dimension is inference,
\texttt{evidence\_status=missing}, and the rationale mentions ``cluster'' or
``robust standard errors''; rule~4, the dimension is concurrent policies and
\texttt{retrieval\_quality=good}. Rules~2--4 are post-hoc string and field
rules defined after inspecting the errors on this same pilot, and all three
cells they touch land on the gold label. They are not principled: naming an
estimator, or the Goodman-Bacon decomposition (a diagnostic of two-way
fixed-effects comparisons, not a heterogeneity-robust estimator), does not show
that it was used appropriately; a rationale that mentions clustering may be
reporting its absence; and rule~4 has no content condition at all (locating
relevant evidence is not the evidence supporting the assumption). We treat only
rule~1 as a result. The script
(\path{calibrate_argus_v1_pilot.py}) reproduces the historical outputs on all 55
cells.

\begin{figure*}[t]
  \centering
  \includegraphics[width=0.94\textwidth]{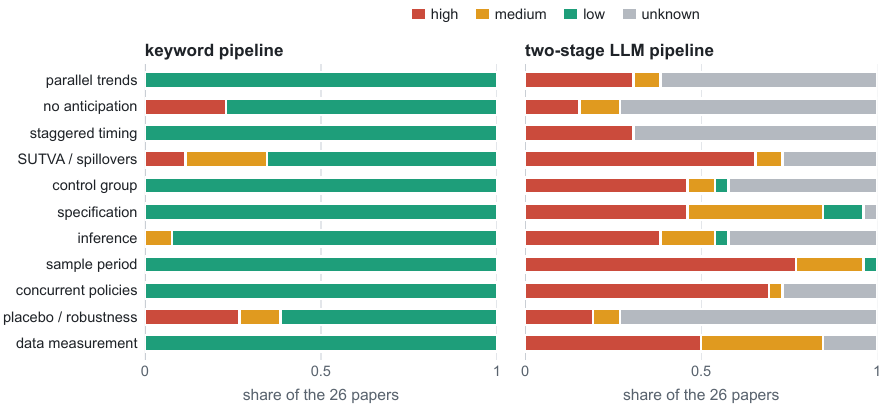}
  \caption{Per-dimension risk distribution on the 26 papers tagged DID in the
  source corpus. The keyword
  baseline (left) is almost entirely \emph{low}; the
  two-stage LLM (right) is differentiated, with a large \texttt{unknown} share
  that marks retrieval failure rather than over-flagging.}
  \label{fig:realcorpus}
\end{figure*}

\begin{figure}[t]
  \centering
  \includegraphics[width=\columnwidth]{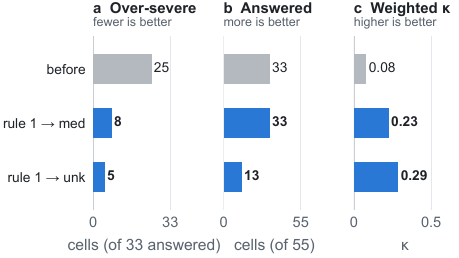}
  \caption{\textbf{Calibration trade-off: severity reduction versus coverage}
  against the reconciled human gold, for the weak-retrieval rule alone
  (in-sample; med and unk are the two policies of Table~\ref{tab:calib}, which
  also gives the exploratory four-rule set).
  Demoting weak-retrieval \emph{high}s to \emph{medium} preserves coverage (33
  answered) while cutting over-severe judgements from 25 to 8; abstaining leaves
  fewer over-severe cells (5) and a higher weighted $\kappa$ ($0.29$) only by
  discarding 20 answers, and leaves the remaining labels unchanged. Most of the
  gain in exact agreement reflects a gold that is $84\%$ \emph{medium}
  (\S\ref{sec:calib}); the remaining bottleneck is evidence grounding, not
  severity thresholding alone.}
  \label{fig:calib_tradeoff}
\end{figure}

The remaining material is Figure~\ref{fig:realcorpus} (per-dimension risk on the
real corpus) and Tables~\ref{tab:vsgold}--\ref{tab:calib} with
Figure~\ref{fig:calib_tradeoff} (agreement and calibration against the gold).

\begin{table}[t]
  \centering\small
  \caption{Two-stage \sys{} vs reconciled human gold (55 cells, 5 papers) under two
  \texttt{unknown} policies: coverage (drop abstentions) and strict
  (\texttt{unknown} counts as mismatch). Pilot-scale: only 2 gold cells are
  \emph{high}; the robust signal is the one-directional over-severity and the
  abstention rate.}
  \label{tab:vsgold}
  \setlength{\tabcolsep}{4pt}
  \begin{tabular}{lcccc}
    \toprule
    \rowcolor{tablehead} \texttt{unknown} policy & $n$ & exact & $\kappa$ & wt.\ $\kappa$ \\
    \midrule
    exclude (coverage) & 33 & 0.242 & 0.031 & 0.128 \\
    mismatch (strict)  & 55 & 0.145 & 0.007 & -- \\
    \bottomrule
  \end{tabular}
\end{table}

\begin{figure*}[t]
  \centering
  \includegraphics[width=0.94\textwidth]{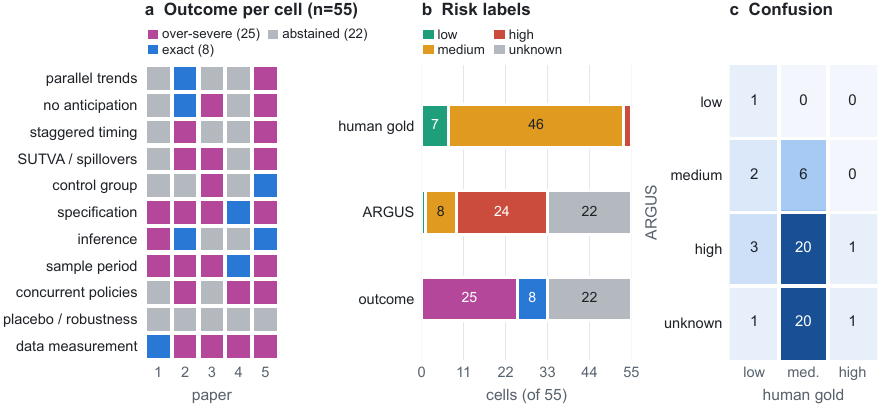}
  \caption{Two-stage \sys{} versus the reconciled gold on the 55 pilot cells
  (frozen protocol). \textbf{(a)}~Every cell, eleven dimensions by five papers,
  coloured by outcome: \sys{} abstains on 22 and, of the 33 it answers, is more
  severe than the gold on 25, matches it on 8, and is less severe on none.
  \textbf{(b)}~Distribution of risk labels in the gold and in \sys{}'s output,
  and the outcome of the 55 cells in the colours of panel (a).
  \textbf{(c)}~\sys{}-by-gold confusion matrix (counts).}
  \label{fig:vsgold}
\end{figure*}

\begin{table}[t]
  \centering\small
  \caption{Deterministic calibration vs the human gold (55 cells), in-sample.
  Two policies for a weak-retrieval \emph{high}: demote to \emph{medium} (med,
  keep coverage) or abstain to \texttt{unknown} (unk). Rule~1 alone is the
  result we claim; rules~1--4 add three single-cell exploratory rules
  (Appendix~\ref{app:supp}). Under abstention rule~1 changes no retained
  label: its higher agreement comes from dropping 20 answers. Pilot scale: only 2 gold cells are
  \emph{high}; a constant \emph{medium} label scores $26/33$ on the answered
  cells, so exact agreement carries little information here. The before column is the rich re-run of \S\ref{sec:calib}, which
  differs from the corpus run of Table~\ref{tab:vsgold} in one cell (a gold-low
  cell moves from \emph{medium} to \emph{high}, which is why weighted $\kappa$
  is 0.08 here and 0.13 there); intervals in Table~\ref{tab:ci_gold}.}
  \label{tab:calib}
  \setlength{\tabcolsep}{4pt}
  \resizebox{\columnwidth}{!}{%
  \begin{tabular}{lccccc}
    \toprule
     \rowcolor{tablehead} & & \multicolumn{2}{c}{rule 1} & \multicolumn{2}{c}{rules 1--4} \\
    \cmidrule(lr){3-4}\cmidrule(lr){5-6}
    \rowcolor{tablehead} metric & before & med & unk & med & unk \\
    \midrule
    over-severe (answ.)     & 25 & 8 & 5 & 5 & 2 \\
    high precision          & 1/25 & 1/5 & 1/5 & 1/2 & 1/2 \\
    high recall             & 1/2 & 1/2 & 1/2 & 1/2 & 1/2 \\
    exact agr.\ (answ.)     & 0.24 & 0.76 & 0.62 & 0.85 & 0.85 \\
    wt.\ $\kappa$ (answ.)   & 0.08 & 0.23 & 0.29 & 0.22 & 0.32 \\
    answered / \texttt{unk} & 33/22 & 33/22 & 13/42 & 33/22 & 13/42 \\
    \bottomrule
  \end{tabular}}
\end{table}

\section{Shared-Evidence Control}
\label{app:shared}

Table~\ref{tab:results} compares pipelines that differ in the judgement policy
\emph{and} in the evidence they retrieve. This control holds the evidence fixed.
Its design, thresholds, and decision rule were committed before this control
was run (\path{SHARED_EVIDENCE_PROTOCOL.md}, commit \texttt{e6dfd70}, 20
September 2026; the runner, the two zero-call cells, and the model calls
followed). Two evidence bundles are
frozen to disk for every paper and dimension: K, the keyword
pipeline's stage-2 chunks, and S, the LLM path's four section candidates with
the relevance gate switched off. Each bundle is read by both policies, the
keyword scorer and the adequacy judge (gpt-4o-2024-11-20, temperature~$0$,
prompts of Appendix~\ref{app:prompts} verbatim). No cell is gated, so none can
abstain. All 45 papers (one clean fixture, 11 flaws, 33 variants) are judged on
all eleven dimensions: 990 adequacy calls, none failed. Cell K-kw reproduces the
published keyword baseline exactly.

\begin{table}[t]
  \centering\small
  \caption{Shared-evidence control: evidence bundle (K, keyword chunks; S,
  ungated sections) $\times$ judgement policy (kw, keyword scorer; llm, adequacy
  judge). det / fa / loc as in Table~\ref{tab:matrix}; off = share of the ten
  non-target dimensions rated $\geq$ medium on the injected variants. The
  reference row is the gated pipeline on the gpt-4o alias (detection and false
  alarm from run~1, Table~\ref{tab:ci}; localization is the three-run mean of
  Table~\ref{tab:variants}).}
  \label{tab:shared}
  \setlength{\tabcolsep}{3pt}
  \resizebox{\columnwidth}{!}{%
  \begin{tabular}{lccccccc}
    \toprule
     \rowcolor{tablehead} & \multicolumn{3}{c}{11 flaws} & \multicolumn{4}{c}{33 variants} \\
    \cmidrule(lr){2-4}\cmidrule(lr){5-8}
    \rowcolor{tablehead} cell & det & fa & loc & det & fa & loc & off \\
    \midrule
    K-kw  & 2/11  & 0/11 & 2/11  & 8/33  & 0/33 & 8/33  & 0.00 \\
    K-llm & 10/11 & 1/11 & 8/11  & 31/33 & 3/33 & 22/33 & 0.15 \\
    S-kw  & 1/11  & 0/11 & 1/11  & 2/33  & 0/33 & 2/33  & 0.00 \\
    S-llm & 11/11 & 0/11 & 11/11 & 31/33 & 0/33 & 22/33 & 0.16 \\
    \midrule
    \textit{gated two-stage} & 8/11 & 0/11 & 8/11 & 25/33 & 3/33 & 0.66 & -- \\
    \bottomrule
  \end{tabular}}
\end{table}

\paragraph{Result.}
With the evidence identical, the policy contrast is large under both bundles
(Table~\ref{tab:shared}): on K, detection rises from 2 to 10 of 11 flaws (eight
gained, none lost; exact McNemar $p{=}0.008$) and from 8 to 31 of 33 variants
($p{<}0.001$); on S, from 1 to 11 and from 2 to 31. The evidence contrast at a
fixed judge is small (10 vs 11 of 11; 31 vs 31 of 33). The pre-specified rule
asked for a gain of at least $0.30$ on the 11 flaws, the same direction on the
variants, and false-alarm and clean off-target rates within $0.10$ of the
keyword scorer's. All three hold, the last narrowly ($0.09$), so the policy
explanation is supported: on these fixtures the judgement policy, not the
retrieval path, carries the gap in Table~\ref{tab:results}.

\paragraph{What it costs, and what it does not show.}
The gain is a trade. The judge rates $11$--$16\%$ of non-target dimensions
medium or higher on injected papers, where the keyword scorer flags none. On
the keyword chunks
the judge also rates one clean dimension (SUTVA) medium. Every false alarm in
the K-llm row is that single verdict, counted once per flaw that targets the
dimension: both benchmarks share one clean fixture, so the false-alarm rates
here and in Tables~\ref{tab:results} and~\ref{tab:variants} rest on one clean
audit of eleven dimensions, not on 11 or 33 independent trials. The ungated
S-llm cell detects more than the gated pipeline (omission variants 11/11 vs
5/11), the trade \S\ref{sec:variants} describes: shown four sections of a short
fixture, the judge notices the deleted one, where the gate abstains. The
reference row was run on the gpt-4o alias, so that comparison is indicative;
fixtures, sentinel lists, and injector are co-designed, and $n$ is small.

\end{document}